\documentclass{sig-alternate-05-2015}
\usepackage{graphicx}
\usepackage{placeins}
\usepackage{wasysym}
\usepackage{amssymb}
\usepackage{amsmath}
\usepackage{array}
\newcolumntype{L}[1]{>{\raggedright\let\newline\\\arraybackslash\hspace{0pt}}m{#1}}
\newcolumntype{C}[1]{>{\centering\let\newline\\\arraybackslash\hspace{0pt}}m{#1}}
\newcolumntype{R}[1]{>{\raggedleft\let\newline\\\arraybackslash\hspace{0pt}}m{#1}}
\usepackage{booktabs}
\usepackage{tabularx}
\usepackage{color}
\usepackage{floatflt}
\usepackage[table]{xcolor}
\definecolor{Gray}{gray}{0.9}
\definecolor{White}{rgb}{1,1,1}
\definecolor{Gray}{gray}{0.9}
\definecolor{LightCyan}{rgb}{0.88,1,1}
\usepackage{soul}
\usepackage{sidecap}
\graphicspath{{./Figures/}}
\usepackage{hyperref}
\hypersetup{colorlinks=true,
citecolor=blue,
linkcolor=black,
urlcolor=black}

\begin{document}
\CopyrightYear{2016} 
\setcopyright{acmlicensed}
\conferenceinfo{GECCO '16,}{July 20 - 24, 2016, Denver, CO, USA}
\isbn{978-1-4503-4206-3/16/07}\acmPrice{\$15.00}
\doi{http://dx.doi.org/10.1145/2908812.2908898}

\clubpenalty = 10000
\widowpenalty = 10000

\title{{\bf \huge $\epsilon$}-Lexicase Selection for Regression}
\numberofauthors{3}
\author{
\alignauthor
William La Cava\titlenote{corresponding author} \\
       \affaddr{Department of Mechanical and Industrial Engineering}\\
       \affaddr{University of Massachusetts}\\
       \affaddr{Amherst, MA 01003}\\
       \email{wlacava@umass.edu}
%%\alignauthor
%%Thomas Helmuth\\
%%       \affaddr{Department of Computer Science}\\
%%       \affaddr{University of Massachusetts}\\
%%       \affaddr{Amherst, MA 01003}\\
%%       \email{thelmuth@umass.edu}
\alignauthor
Lee Spector\\
       \affaddr{School of Cognitive Science}\\
       \affaddr{Hampshire College}\\
       \affaddr{Amherst, MA 01002}\\
       \email{lspector@hampshire.edu}
\and
\alignauthor
Kourosh Danai\\
       \affaddr{Department of Mechanical and Industrial Engineering}\\
       \affaddr{University of Massachusetts}\\
       \affaddr{Amherst, MA 01003}\\
       \email{danai@engin.umass.edu}
}

\maketitle
%\ead{danai@ecs.umass.edu}
%\address[mie]{D, , Amherst, MA 01003}
%\address[hc]{School of Cognitive Science, Hampshire College, Amherst MA 01002}
%
\begin{abstract}
Lexicase selection is a parent selection method that considers test cases separately, rather than in aggregate, when performing parent selection. It performs well in discrete error spaces but not on the continuous-valued problems that compose most system identification tasks. In this paper, we develop a new form of lexicase selection for symbolic regression, named $\epsilon$-lexicase selection, that redefines the pass condition for individuals on each test case in a more effective way. We run a series of experiments on real-world and synthetic problems with several treatments of $\epsilon$ and quantify how $\epsilon$ affects parent selection and model performance. $\epsilon$-lexicase selection is shown to be effective for regression, producing better fit models compared to other techniques such as tournament selection and age-fitness Pareto optimization. We demonstrate that $\epsilon$ can be adapted automatically for individual test cases based on the population performance distribution. Our experiments show that $\epsilon$-lexicase selection with automatic $\epsilon$ produces the most accurate models across tested problems with negligible computational overhead. We show that behavioral diversity is exceptionally high in lexicase selection treatments, and that $\epsilon$-lexicase selection makes use of more fitness cases when selecting parents than lexicase selection, which helps explain the performance improvement.\footnote{\hl{Note:} this is a corrected version of the original GECCO '16 conference paper. Equations 2 and 5 have been corrected to indicate that the pass conditions for individuals in $\epsilon$-lexicase selection are defined relative to the best error in the population on that training case, not in the selection pool.}
\end{abstract}

%\category{I.2.2}{Artificial Intelligence}{Automatic Programming}[Program synthesis]
%\category{J.2}{Computer Applications}{Physical Science and Engineering}[Engineering]
%\terms{Algorithms}


%\ccsdesc[500]{Mathematics of computing~Evolutionary algorithms}

%\printccsdesc


\section{INTRODUCTION}
Genetic programming (GP) traditionally tests programs on many test cases and then reduces the performance into a single value that is used to select parents for the next generation. Typically the fitness $f$ of an individual is quantified as its aggregate performance over the training set $\mathcal{T} = \{ (y_t,\mathbf{x}_t)\}_{t = 1}^N$, using e.g. the mean absolute error (MAE), which is quantified for individual program $i \in P$ as:
\begin{equation}
f(i,\mathcal{T}) = \frac{1}{N} \sum_{t \in \mathcal{T}}{|y_t - \hat{y}_t(i,\mathbf{x}_t)|} \label{eq:fit} 
\end{equation}
where $\mathbf{x} \in \mathbb{R}^D$ represents the variables or features, the target output is $y$ and $\hat{y}(i,\mathbf{x})$ is the program's output. As a result of the aggregation of the absolute error vector $e(i) = |y - \hat{y}(i,\mathbf{x})|$ in Eq.~(\ref{eq:fit}), the relationship of $\hat{y}$ to $y$ is represented crudely when choosing models to propagate. As others have pointed out~\cite{krawiec_automatic_2015}, aggregate fitnesses strongly reduce the information conveyed to GP about $i$ relative to the description of $i's$ behavior available in $e(i)$, thereby under-utilizing information that could help guide the search. In addition, many forms of aggregation assume all tests are equally informative (although there are exceptions, including implicit fitness sharing which is discussed below). Therefore individuals that are elite (i.e. have the lowest error in the population) for portions of $e$ are not selected if they perform poorly in other regions and therefore have a higher $f$. By providing equivalent selection pressure with respect to test cases, GP misses the opportunity to identify programs that perform especially well in certain regions of the problem, most importantly those portions of the problem that are more difficult for the process to solve. We expect GP to solve problems through the induction, propagation and recombination of building blocks (i.e. subprograms) that provide partial solutions to our desired task. Hence we wish to select those programs that imply a partial solution by performing uniquely well on subsets of the problem.  

Several methods have been proposed to reward individuals with uniquely good test performance, such as implicit fitness sharing (IFS)~\cite{mckay_investigation_2001}, historically assessed hardness~\cite{klein_genetic_2008}, and co-solvability~\cite{schaefer_using_2010}, all of which assign greater weight to fitness cases that are judged to be more difficult in view of the population performance. Perhaps the most effective parent selection method recently proposed is lexicase selection~\cite{helmuth_solving_2014,spector_assessment_2012}. In particular, ``global  pool,  uniform  random  sequence,  elitist  lexicase  selection"~\cite{spector_assessment_2012}, which we refer to simply as lexicase selection, has outperformed other similarly-motivated methods in recent studies~\cite{helmuth_general_2015-1,liskowski_comparison_2015}. Despite these gains, it fails to produce such benefits when applied to continuous symbolic regression problems, due to its method of selecting individuals based on test case elitism. We demonstrate in this paper that by re-defining the test case pass condition in lexicase selection using an $\epsilon$ threshold, the benefits of lexicase selection can be achieved in continuous domains. 

We begin by describing the $\epsilon$-lexicase selection algorithm in \S\ref{s:2} and discuss how it differs with respect to standard lexicase selection. Several definitions of $\epsilon$ are proposed. We briefly review related work in \S\ref{s:3} and describe the relation between lexicase selection and multiobjective methods. The experimental analysis is presented in \S\ref{s:4}, beginning with a parameter variation study of $\epsilon$ and ending with a comparison of several GP methods on a set of real-world and symbolic regression problems. Given the results, we propose future research directions in \S\ref{s:5} and summarize our findings in \S\ref{s:6}.   

\section{{\large $\epsilon$} Lexicase Selection}\label{s:2}
Lexicase selection is a parent selection technique based on lexicographic ordering of test (i.e. fitness) cases.  Each parent selection event proceeds as follows: 
\begin{enumerate}
\item The entire population is added to the selection pool.
\item The fitness cases are shuffled.
\item Individuals in the pool with a fitness worse than the best fitness on this case among the pool are removed. 
\item If more than one individual remains in the pool, the first case is removed and 3 is repeated with the next case. If only one individual remains, it is the chosen parent. If no more fitness cases are left, a parent is chosen randomly from the remaining individuals. 
\end{enumerate}  
As evidenced above, the algorithm is quite simple to implement. In this procedure, test cases act as filters, and a randomized path through these filters is constructed each time a parent is selected. Each parent selection event returns a parent that is elite on at least the first test case used to select it. In turn, the filtering capacity of a test case is directly proportional to its difficulty since it culls the individuals from the pool that do not do the best on it. Therefore selective pressure continually shifts to individuals that are elite on cases that are not widely solved in the population. Because each parent is selected via a randomized ordering of test cases and these cases perform filtering proportional to their difficulty, individuals are pressured to perform well on unique combinations of test cases, which promotes individuals with diverse performance, leading to increased diversity observed during evolutionary runs~\cite{helmuth_solving_2014}. 

Lexicase selection was originally applied to multimodal~\cite{spector_assessment_2012} and ``uncompromising"~\cite{helmuth_solving_2014} problems. An uncompromising problem is one in which only exact solutions to every test case produce a satisfactory program. For those types of problems, using each case as a way to select only elite individuals is well-motivated, since each test case must be solved exactly. In regression, exact solutions to test cases can only be expected for synthetic problems, whereas real-world problems are subject to noise and measurement error. With respect to the lexicase selection process, continuously-valued errors are problematic, due to the fact that individuals in the population are not likely to share elitism on any particular case unless they are identical equations. On regression problems, the standard lexicase procedure typically uses only one case for each parent selection, resulting in poor performance. 

We hypothesize that lexicase selection performs poorly on continuous errors because the case passing criteria is too stringent in continuous error spaces. For individual $i$ to pass case $t$, lexicase requires that $e_t(i) = e^*_t$, where $e^*_t$ is the best error on that test case in the pool. To remedy this shortcoming, we introduce $\epsilon$-lexicase selection, which modulates the pass condition on test cases via a parameter $\epsilon$, such that only individuals outside of a predefined $\epsilon$ are filtered in step 3 of lexicase selection. We experiment with four different definitions of $\epsilon$ in this paper. The first two, $\epsilon_e$ and $\epsilon_y$, are absolute thresholds that define the pass condition $p_t(i)$ of program $i$ on test case $t$ as follows:

\begin{align}
\epsilon_e &: p_t(i) = \mathbb{I}\left(e_t(i) <  \mathbf{e}^*_t (1+\epsilon_e) \right) \label{eq:ep_e} \\ 
\epsilon_y &: p_t(i) = \mathbb{I}\left(e_t(i) < \epsilon_y \right) \label{eq:ep_y}
\end{align}
Here $\mathbb{I}$ is the indicator function that returns 1 if true and 0 if false, \hl{and $\mathbf{e}^*_t$ is the best error on case $t$ in $P$}. As shown in Eq.~(\ref{eq:ep_e}), $\epsilon_e$ defines $p_t(i)$ relative to $\mathbf{e}^*_t$, and therefore is always passed by at least one individual in $P$. Conversely, $\epsilon_y$ (Eq.~(\ref{eq:ep_y})) defines $p_t(i)$ relative to the target value $y_t$, meaning that $\hat{y}_t(i)$ must be within $\pm \epsilon_y$ of $y_t$ to pass case $t$. In this way $\epsilon_y$ provides no selection pressure if there is not an individual in the population within adequate range of the true value for that case. 

$\epsilon_e$ and $\epsilon_y$ are the simplest definitions of $\epsilon$-lexicase selection, but have two distinct disadvantages: 1) they have to be specified by the user, and 2) their optimal values are problem dependent. An absolute $\epsilon$ is unable to provide a desired amount of filtering in each selection event since it is blind to the population's performance. Ideally $\epsilon$ should automatically adapt to take into account the values of $e_t(i)$ across $P$, denoted $\mathbf{e}_t \in \mathbb{R}^{|P|}$, so that it can modulate its selectivity based on the difficulty of $t$. A common estimate of difficulty in performance on a fitness case is variance~\cite{schmidt_coevolution_2008}; in this regard $\epsilon$ could be defined according to the standard deviation of $\mathbf{e}_t$, i.e. $\sigma(\mathbf{e}_t)$. Given the high sensitivity of $\sigma$ to outliers, however, we opt for a more robust estimation of variability by using the median absolute deviation (MAD)~\cite{pham-gia_mean_2001} of $\mathbf{e}_t$, defined as
\begin{equation}
MAD(\mathbf{e}_t) = \lambda(\mathbf{e}_t) = \text{median}_j \left( |\mathbf{e}_{t_j} - \text{median}_k(\mathbf{e}_{t_k})| \right) \label{eq:mad}
\end{equation}
We use Eq.~(\ref{eq:mad}) in the definition of two $\epsilon$ values, $\epsilon_{e \lambda}$ and $\epsilon_{y \lambda}$, that are defined analogously to $\epsilon_e$ and $\epsilon_y$ as:
\begin{align}
\epsilon_{e \lambda} &: p_t(i) = \mathbb{I}\left(e_t(i) < \mathbf{e}^*_t + \lambda(\mathbf{e}_t) \right) \label{eq:ep_emad}\\ 
\epsilon_{y \lambda} &: p_t(i) = \mathbb{I}\left(e_t(i) <  \lambda(\mathbf{e}_t) \right)\label{eq:ep_ymad}
\end{align}

An important consideration in parent selection is the time complexity of the selection procedure. Lexicase selection has a theoretical worst-case time complexity of $O(|P|^2N)$, compared to a time complexity of $O(|P|N)$ for tournament selection. Although clearly undesirable, this worst-case complexity is only reached if every individual passes every test case during selection; in practice~\cite{helmuth_solving_2014}, lexicase selection normally uses a small number of cases for each selection and therefore incurs only a small amount of overhead. We quantify the wall clock times for our variants of lexicase compared to other methods in \S\ref{s:4 results}.  
\section{Related Work}\label{s:3}
Although to an extent the ideas of multiobjective optimization apply to multiple test cases, they are qualitatively different: objectives are the defined goals of a task, whereas test cases are tools for estimating progress towards those objectives. Objectives and test cases therefore commonly exist at different scales: symbolic regression often involves one or two objectives (e.g. accuracy and model conciseness) and hundreds or thousands of test cases. One example of using test cases explicitly as objectives occurs in Langdon's work on data structures~\cite{langdon_evolving_1995} in which small numbers of test cases (in this case 6) are used as multiple objectives in a Pareto selection scheme. Other multi-objective approaches such as NSGA-II~\cite{schoenauer_fast_2000}, SPEA2~\cite{zitzler_spea2:_2001} and ParetoGP~\cite{smits_pareto-front_2005} are used commonly with a small set of objectives in symbolic regression. The ``curse of dimensionality" prevents the use of objectives at the scale of typical test case sizes, since most individuals become nondominated\footnote{Program $i_1$ dominates $i_2$ if $f_j({i}_1) \leq f_j({i}_2)\;\forall j$ and $f_j({i}_1) < f_j({i}_2)$ for at least one $j$ ($f$ is minimized).}, leading to selection based mostly on expensive diversity measures rather than performance. Scaling issues in many-objective optimization are reviewed in~\cite{ishibuchi_evolutionary_2008}. In lexicase selection, parents are guaranteed to be nondominated with respect to the fitness cases. Pareto strength in SPEA2 promotes individuals based on how many individuals they dominate, and similarly lexicase selection increases the probability of selection for individuals who solve {\it more} cases and {\it harder} cases (i.e. cases that are not solved by other individuals) and decreases for individuals who solve {\it fewer} or {\it easier} cases. 

A number of GP methods attempt to affect selection by weighting test cases based on population performance. In non-binary Implicit Fitness Sharing (IFS)~\cite{krawiec_implicit_2013}, the fitness proportion of a case is scaled by the performance of other individuals on that case. Similarly, historically assessed hardness scales error on each test case by the success rate of the population~\cite{klein_genetic_2008}. Discovery of objectives by clustering (DOC)~\cite{krawiec_automatic_2015} clusters test cases by population performance, and thereby reduces test cases into a set of objectives for search. Both IFS and DOC were outperformed by lexicase selection on program synthesis and boolean problems in previous studies~\cite{helmuth_general_2015-1,liskowski_comparison_2015}. Other methods attempt to sample a subset of $\mathcal{T}$ to reduce computation time or improve performance, such as dynamic subset selection~\cite{gathercole_dynamic_1994}, interleaved sampling~\cite{goncalves_balancing_2013}, and co-evolved fitness predictors~\cite{schmidt_coevolution_2008}. Unlike these methods, lexicase selection begins each selection with the full set of training cases, and allows selection to adapt to program performance on them.
 
The conversion of a model's real-valued fitness into discrete values based on an $\epsilon$ threshold has been explored in other research; for example, Novelty Search GP~\cite{martinez_searching_2013} uses a reduced error vector to define behavioral representation of individuals in the population. This paper proposes it for the first time as a solution to applying lexicase selection effectively to regression.

As a behavioral-based search driver, lexicase selection belongs to a class of GP systems that attempt to incorporate a program's behavior explicitly into the search process, and as such shares a general motivation with recently proposed methods such as Semantic GP~\cite{moraglio_geometric_2012} and Behavioral GP~\cite{krawiec_behavioral_2014}, despite differing strongly in approach. Although lexicase is designed with behavioral diversity in mind, recent studies suggest that structural diversity can also significantly affect GP performance~\cite{burks_efficient_2015}.

\section{Experimental Analysis} \label{s:4}
We define the problems used to assess $\epsilon$-lexicase selection here, as well as a set of existing GP methods used for comparison. We then analyze and tune the value of $\epsilon_e$ and $\epsilon_y$ on an example problem and discuss the results. Finally we test all of the methods on each problem and summarize the findings.  
\subsection{Problems}
Three synthetic and three real-world problems were chosen for benchmarking different GP methods. The first problem is the housing data set~\cite{harrison_hedonic_1978} that seeks a model to estimate Boston housing prices. The second problem is the Tower problem\footnote{http://symbolicregression.com/?q=towerProblem} that consists of 15-minute averaged time series data taken from a chemical distillation tower, with the goal of predicting propelyne concentration. The third problem, referred to as the Wind problem~\cite{la_cava_automatic_2016}, features data collected from the Controls and Advanced Research Turbine, a 600 kW wind turbine operated by the National Wind Technology Center. The data set consists of time-series measurements of wind speed, control actions, and acceleration measurements that are used to predict the bending moment measured at the base of the wind turbine. In this case solutions are formulated as first-order discrete-time dynamic models of the form $\hat{y} = f(\mathbf{x}, \mathbf{x}_{t-1}, {\hat{y}}_{t-1})$. The fourth and fifth problem tasks are to estimate the energy efficiency of heating (ENH) and cooling (ENC) requirements for various simulated buildings~\cite{tsanas_accurate_2012}. The last problem is the UBall5D problem\footnote{UBall5D is also known as Vladislavleva-4.} which has the form \[ y = \frac{10}{5+\sum_{i=1}^5{(x_i-3)^2}}\] The Tower problem and UBall5D were chosen from the benchmark suite suggested by White et. al. \cite{white_better_2012}. The dimensions of all data sets are shown in Table~\ref{tbl:symreg_settings}. Aside from UBall5D which has a pre-defined test set~\cite{vladislavleva_order_2009}, the problems were divided 70/30 into training and testing sets. These sets were normalized to zero mean, unit variance and randomly partitioned for each trial. 

%\paragraph{Evolutionary Scheme}
\subsection{Compared Methods}
Our definitions of $\epsilon$ in \S\ref{s:2} yield four methods which we analyze in our experiments, abbreviated as \text{Lex} \text{$\epsilon_e$}, \text{Lex} \text{$\epsilon_y$}, \text{Lex} \text{$\epsilon_{e \lambda}$}, \text{Lex} \text{$\epsilon_{y \lambda}$}. We compare these variants to standard lexicase selection (denoted as simply \text{Lex}) and standard tournament selection of size 2 (denoted \text{Tourn}). To control for the effect of selection in GP, we also  compare these methods to random parent selection, denoted \text{Rand Sel}.

In addition to these methods, many state-of-the-art symbolic regression tools leverage Pareto optimization \cite{smits_pareto-front_2005, schmidt_distilling_2009,burks_efficient_2015} and/or age layering~\cite{hornby_alps:_2006} to improve symbolic regression performance. With this in mind, we also compare $\epsilon$-lexicase selection to age-fitness Pareto survival (\text{AFP}) \cite{schmidt_age-fitness_2011}, in which each individual is assigned an age equal to the number of generations since its oldest ancestor was created. Each generation, a new individual is introduced to the population as a means of random restart. Selection for breeding is random, and during breeding a number of children are created equal to the overall population size. Survival is conducted according to the environmental selection algorithm in SPEA2~\cite{zitzler_spea2:_2001}, as in~\cite{schmidt_machine_2011}.  

Every method uses sub-tree crossover and point mutation as search operators. For each method, we include a parameter hill climbing step each generation that perturbs the constants in each equation with Gaussian noise and saves those changes that improve the model's MAE (Eq.~(\ref{eq:fit})). The complete code for these tests is available online\footnote{\url{https://www.github.com/lacava/ellen}}.
%\paragraph{Crossover and Mutation}

%\\
%R^2_{i} &=& \frac{\left( \text{cov}(y^*,y_{i}) \right)^2}{\text{var}(y^*)  \text{var}(y_{i})} \label{eq:R2}
%\end{eqnarray}

\begin{table}
\scriptsize
\caption{Symbolic regression problem settings.}\label{tbl:symreg_settings}
\begin{tabularx}{\columnwidth}{X R{0.57\columnwidth}} \toprule
Setting& Value \\ \midrule
Population size & 1000 \\
Crossover / mutation & 80/20\% \\
Program length limits & [3, 50] \\ 
ERC range & [-1,1] \\
Generation limit & 1000 \\
Trials & 30 \\
Terminal Set & \{$\mathbf{x}$, ERC, $+$, $-$, $*$, $/$, $\sin$, $\cos$, $\exp$, $\log$\}\\
Elitism & keep best \\
\end{tabularx}
\begin{tabularx}{\columnwidth}{X r r r} \midrule
%\begin{tabular}{l r r r} \hline
Problem & Dimension & Training Cases & Test Cases \\ \midrule
Housing & 14 & 354 & 152 \\
Tower & 25 & 2195 &  940 \\
Wind & 6 & 4200 & 1800 \\
ENH & 8 & 538 & 230 \\
ENC & 8 & 538 & 230 \\
UBall5D & 5 & 1024 & 5000 \\ \bottomrule
\end{tabularx}

\end{table}

\subsection{Parameter tuning}
In the cases of $\epsilon_e$ and $\epsilon_y$, the user must specify fixed parameter values. For both cases we tested the set of parameter values $\{$0.01, 0.05, 0.10, 0.50, 1.0, 5.0, 10.0$\}$ over 15 trials. For $\epsilon_e$, these values mean that an individual's $e_t(i)$ must be within 1\% to 1000\% of $e^*_t$ to pass test $t$. For $\epsilon_y$, $\hat{y}_t(i)$ must be within that range of $y_t$. The parameter study was conducted on the Tower symbolic regression problem, the details of which are shown in Table~\ref{tbl:symreg_settings}. It is important to note that the optimal value of these parameters is problem-dependent, although the best values from the parameter tuning experiment were used for all problems in the subsequent sections. 

\begin{figure}
\centering
  \includegraphics[height = 0.25\textheight]{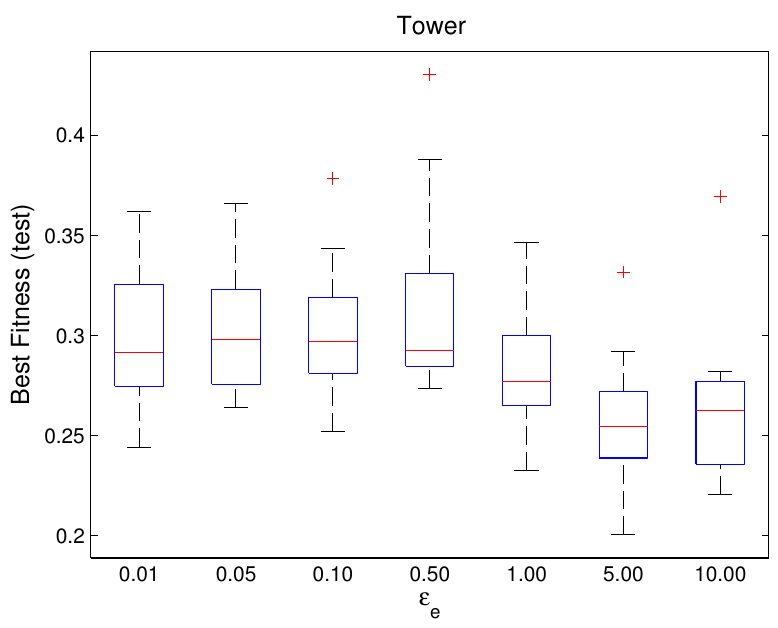}\\
  \caption{Best-of-run fitness on the test set for various levels of $\epsilon_e$.}\label{fig:boxplot_eps_e}
\end{figure}

\begin{figure}
\centering
  \includegraphics[height = 0.25\textheight]{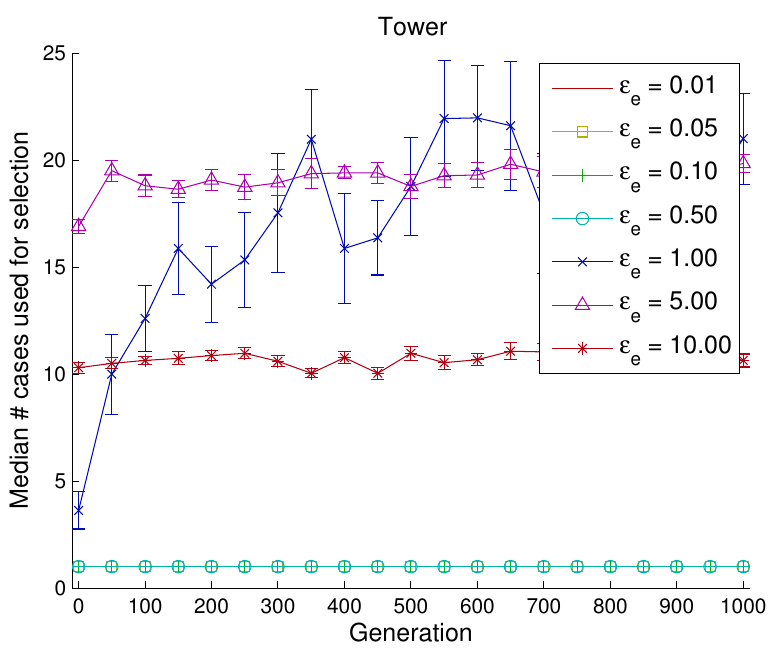}\\
  \caption{Number of test cases used for selection for various levels of $\epsilon_e$. All cases of $\epsilon_e \leq 0.50$ have a median of one case each generation.}\label{fig:lex_cases_eps_e}
\end{figure}

\begin{figure}
\centering
  \includegraphics[height = 0.25\textheight]{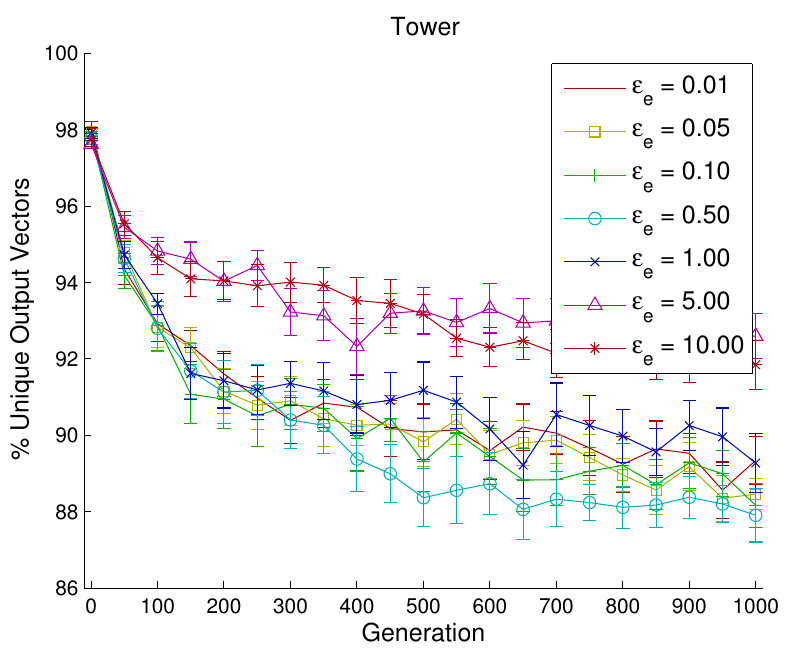}\\
  \caption{Unique output vectors for various levels of $\epsilon_e$.}\label{fig:novelty_eps_e}
\end{figure}

%\subsubsection{\large \bf $\epsilon_e$} 
The test fitness results for different values of $\epsilon_e$ are shown in Figure~\ref{fig:boxplot_eps_e}. The best results are obtained for $\epsilon_e$ = 5.0. The number of cases used during selection are shown in Figure~\ref{fig:lex_cases_eps_e}. This figure matches our intuition about the sensitivity of case usage to $\epsilon_e$: larger tolerances for error use more cases in each selection event. For $\epsilon_e = 1.0$, we observe steady growth in case usage, suggesting population convergence. The diversity of the population's behavior also grows with $\epsilon_e$, as shown by the unique output vectors, i.e. unique $\hat{y}$, plot in Figure~\ref{fig:novelty_eps_e}. For subsequent experiments we use $\epsilon_e = 5.0$ which corresponds to the lowest median test fitness for the Tower problem. 

\begin{figure}
\centering
  \includegraphics[height = 0.25\textheight]{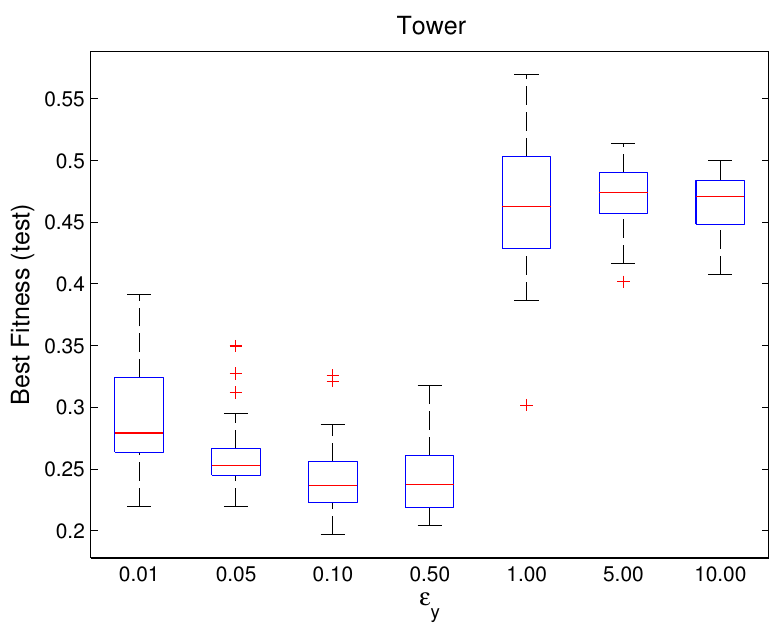}\\
  \caption{Best-of-run fitness the test sets for various levels of $\epsilon_y$.}\label{fig:boxplot_eps_y}
\end{figure}

%\subsubsection{\large \bf $\epsilon_y$} 
The test fitness results for different values of $\epsilon_y$ are shown in Figure~\ref{fig:boxplot_eps_y}. From $\epsilon_y = 1.0$ and upward, we note that \text{Lex} $\epsilon_y$ uses all fitness cases for nearly every selection event, causing long run-times and suggesting that selection has become random. As we show in \S\ref{s:4 results}, Rand Sel performs similarly to $\epsilon_y > 0.50$ on this problem, further supporting the idea of selection pressure loss. We set $\epsilon_y = 0.10$ for the subsequent experiments, again corresponding to the lowest median test fitness. 

%\begin{figure}[htb]
%\centering
%  \includegraphics[width=0.75\columnwidth]{figures/lex_cases_tower_lex_tar_tuning.pdf}\\
%  \caption{Number of test cases used for selection for various levels of $\epsilon_y$.}\label{fig:lex_cases_eps_y}
%\end{figure}
%\begin{figure}[htb]
%\centering
%  \includegraphics[width=0.75\columnwidth]{figures/novelty_tower_lex_tar_tuning.pdf}\\
%  \caption{Unique output vectors for various levels of $\epsilon_y$.}\label{fig:novelty_eps_y}
%\end{figure}

\subsection{Results}\label{s:4 results}

\begin{figure*}
\centering
  \includegraphics[width=\textwidth]{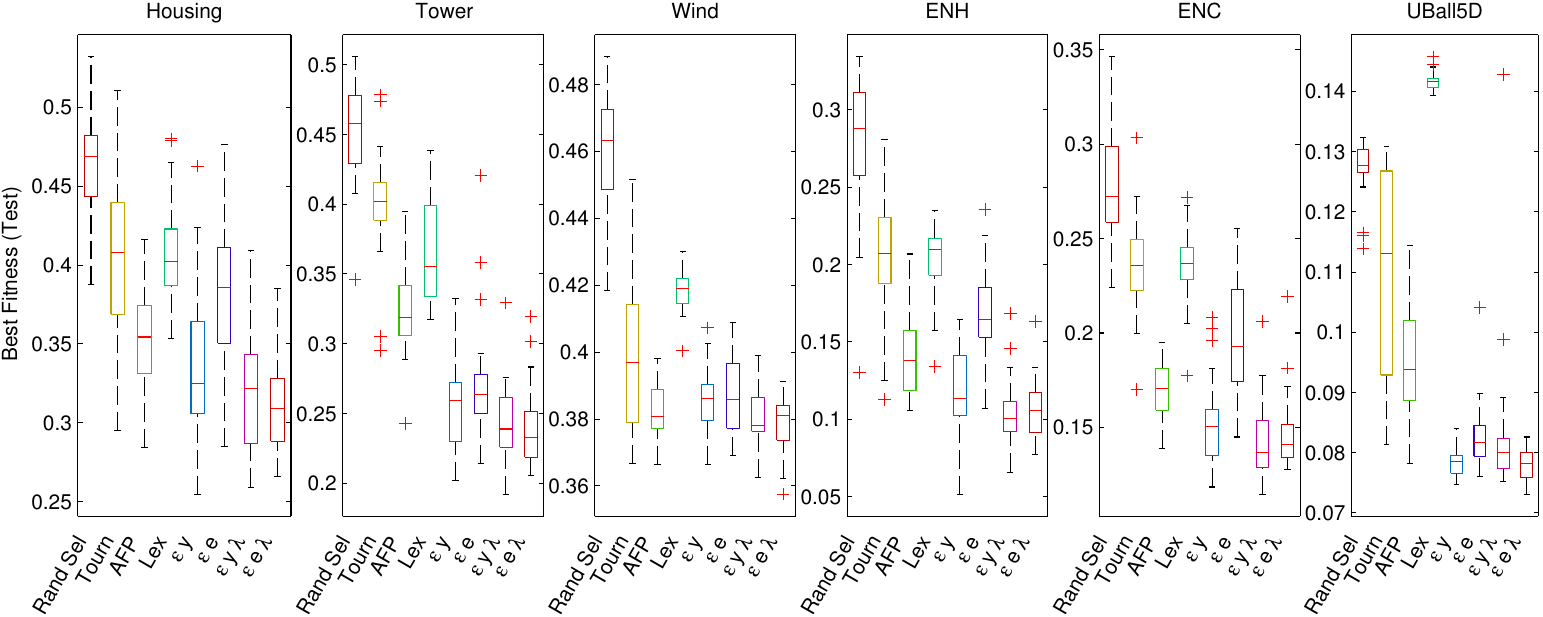}\\
  \caption{Best-of-run fitness statistics on the test sets for all problems.}\label{fig:boxplot_v_all}
\end{figure*}

\begin{figure*}
\centering
  \includegraphics[width=\textwidth]{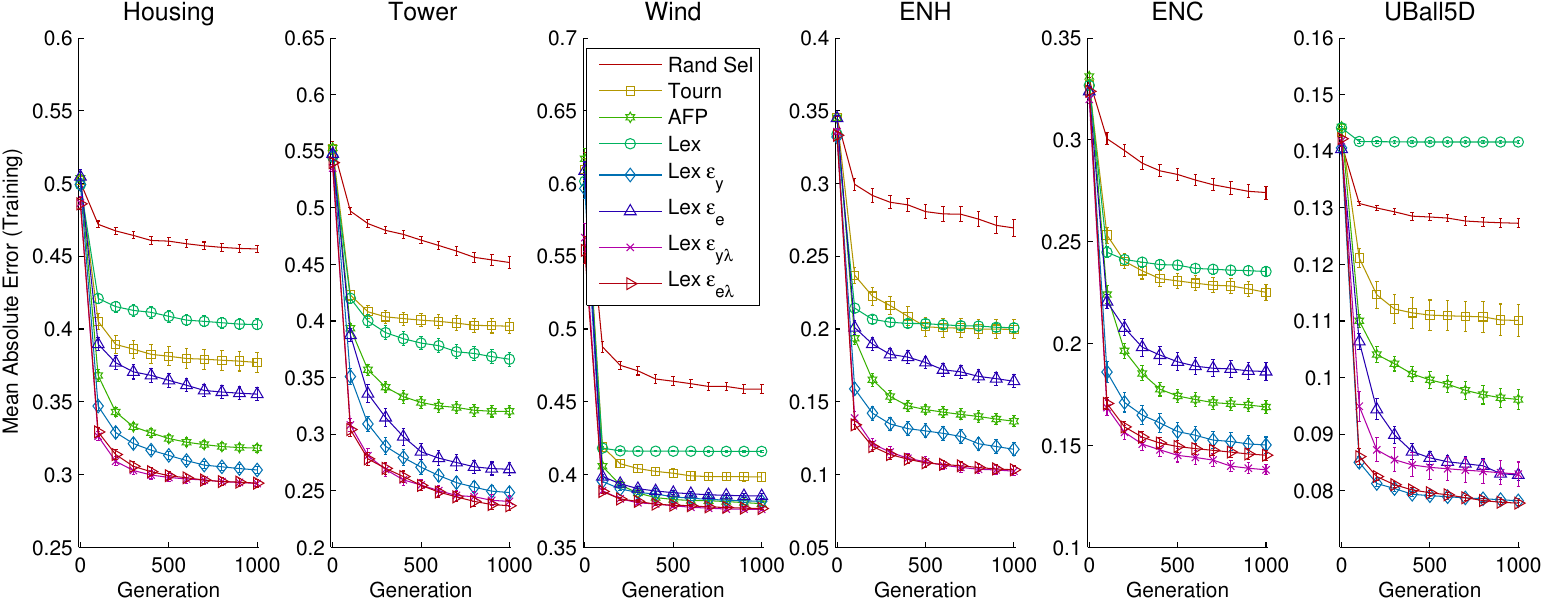}\\
  \caption{Best-of-run fitness each generation on the training sets for all problems. The bars indicate the standard error across all runs.}\label{fig:mae_all}
\end{figure*}

We summarize the experimental results in Table~\ref{tbl:bf_v}. The median best fitness of all runs on the test sets, the mean ranking of each method across problems, and the total run-time to conduct the trials for each method is shown. Figure~\ref{fig:boxplot_v_all} shows the distributions of MAE (Eq.~(\ref{eq:fit})) on the test sets for the best-of-run models. We also visualize the best fitness on the training set each generation in Figure~\ref{fig:mae_all}. 

\begin{figure}
\centering
  \includegraphics[height = 0.25\textheight]{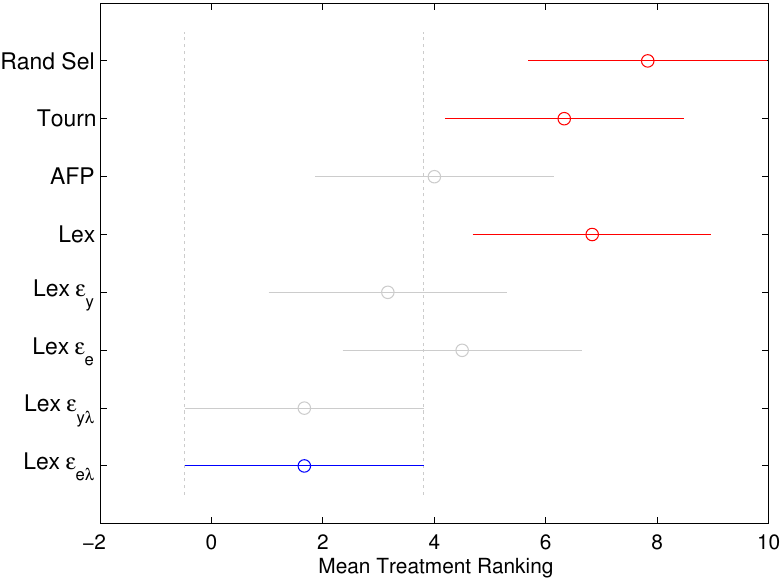}\\
  \caption{Multiple comparison of the Friedman test of mean rankings across all problems, with approximate intervals of significance. Two treatments being compared are significantly different if their intervals do not overlap. Signficant differences from \text{Lex} \text{$\epsilon_{e \lambda}$} (shown in blue) are denoted in red.}\label{fig:friedman}
\end{figure}

Figure~\ref{fig:mae_all} shows that most of the differences in learning on the training set occurs in the first 250 generations, although in all cases \text{Lex} \text{$\epsilon_{e \lambda}$} or \text{$\epsilon_{y \lambda}$} maintains the lowest final training set error. Across all problems, the median best fitness on the test sets is obtained by either \text{Lex} \text{$\epsilon_{e \lambda}$} or \text{Lex} \text{$\epsilon_{y \lambda}$}. According to the pairwise tests annotated in Table~\ref{tbl:bf_v} , \text{Lex} \text{$\epsilon_{e \lambda}$} or \text{Lex} \text{$\epsilon_{y \lambda}$} perform significantly better than \text{Rand Sel}, \text{Tourn}, \text{Lex} and \text{Lex} \text{$\epsilon_e$} on 6/6 problems, better than \text{AFP} on 5/6 problems, and better than \text{Lex} \text{$\epsilon_y$} on one problem. In terms of the mean ranking across tests (Table~\ref{tbl:bf_v}), \text{Lex} \text{$\epsilon_{e \lambda}$} and \text{Lex} \text{$\epsilon_{y \lambda}$} rank the best, followed by \text{Lex} \text{$\epsilon_y$}, \text{AFP}, \text{Lex} \text{$\epsilon_e$}, \text{Tourn}, \text{Lex}, and \text{Rand Sel}, in that order. We conduct a Friedman's test of the mean rankings across problems, the intervals of which are shown in Figure~\ref{fig:friedman}. This comparison indicates that the performance improvement of  \text{Lex} \text{$\epsilon_{e \lambda}$} and  \text{Lex} \text{$\epsilon_{y \lambda}$} relative to \text{Tourn}, \text{Lex}, and \text{Rand Sel} is significant across all tested problems. The intervals show partial overlap with respect to \text{AFP} and \text{Lex} \text{$\epsilon_e$} that may warrant further experiments. 

The median total trial times reported in Table~\ref{tbl:bf_v} indicate that $\epsilon$-lexicase selection takes nearly the same time to finish as tournament selection in practice, despite its higher theoretical worst-case time complexity. On average, \text{Lex} \text{$\epsilon_y$}, $\epsilon_{e}$, \text{$\epsilon_{y \lambda}$} and \text{$\epsilon_{e \lambda}$} report wall clock times that are 96\%, 82\%, 120\%, and 120\% the duration of tournament selection, respectively, giving a negligible average of 105\%. \text{Rand Sel} finishes the fastest due to no selection, and \text{AFP} finishes the slowest, most likely due to the overhead of computing dominance relations for $P$ and, in the case of non-dominated populations, densities in objective space~\cite{zitzler_spea2:_2001}. It is possible that the tournament selection-based version of \text{AFP}~\cite{schmidt_age-fitness_2011} would have a lower run-time, although it may have difficulty scaling to more than two objectives~\cite{schmidt_machine_2011}.     

\begin{figure}
\centering
  \includegraphics[width = \columnwidth]{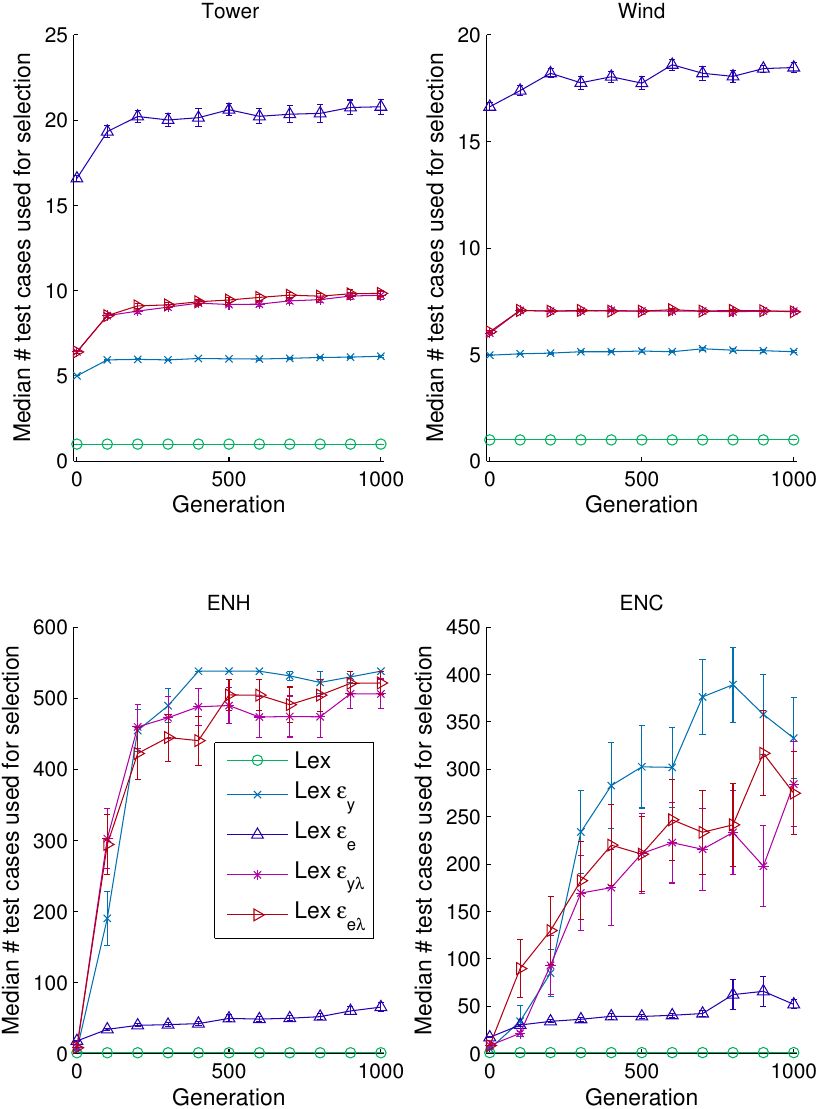}\\
  \caption{Number of fitness cases used in selection for different lexicase methods. The bars indicate the standard error.}\label{fig:lex_cases_all}
\end{figure}

\text{Lex} takes only 41\% of the time of tournament selection to finish, which is explained by its case usage. The number of fitness cases used by lexicase selection variants for four of the problems is shown in Figure~\ref{fig:lex_cases_all}. Note that \text{Lex} uses only one test case during parent selection due to the rarity of elitism in continuous error space; the fact that parents are chosen based on single cases also explains its poor performance. On the Tower and Wind problems, $\epsilon$-lexicase variants show small increases in case usage over the course of evolution. \text{Lex} \text{$\epsilon_e$} uses the highest number of cases on Tower and Wind, but the lowest number of cases for ENH and ENC among $\epsilon$-lexicase variants. On ENH and ENC, a higher percent of total cases are used during selection compared to Tower and Wind, indicating the problem-dependent nature of GP population structures and performance. \text{Lex} \text{$\epsilon_{y \lambda}$} and \text{Lex} \text{$\epsilon_{e \lambda}$} use nearly the same numbers of cases for selection on each problem, which suggests that the performance of $\lambda$-based $\epsilon$  is robust to being defined relative to $e$ or $y$ (Eq.~(\ref{eq:ep_emad}) and~(\ref{eq:ep_ymad})). \text{Lex} \text{$\epsilon_e$} and \text{$\epsilon_y$}, on the other hand, vary strongly across problems in terms of their case usage, again indicating their parametric sensitivity. 

\begin{figure}
\centering
  \includegraphics[width = \columnwidth]{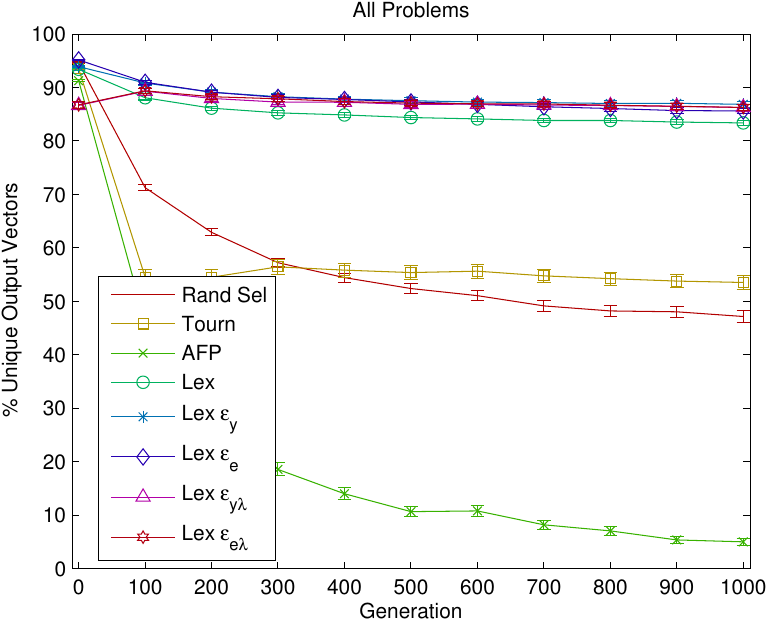}\\
  \caption{Mean population diversity as a function of generation for each method over all problems. The bars indicate the standard error.}\label{fig:novelty_summary}
\end{figure}

We observe exceptionally high population diversity for the lexicase methods, which supports observations in~\cite{helmuth_solving_2014}. We measure diversity by the percent of unique $\hat{y}$ among programs in $P$, plotted as an average across problems in Figure~\ref{fig:novelty_summary}. Interestingly, the diversity is higher using lexicase than random selection, which indicates lexicase selection's ability to exploit behavioral difference to increase diversity beyond the search operator effects. The differential performance between \text{Rand Sel} and $\epsilon$-lexicase selection shown in Table~\ref{tbl:bf_v} demonstrates that the gains afforded by $\epsilon$-lexicase selection are not due to simply increased randomization, but rather the promotion of individuals with exceptionally good performance on diverse orderings of test cases.

\begin{table*}
\scriptsize
\centering
\renewcommand{\arraystretch}{1.2}\addtolength{\tabcolsep}{.5pt}
\caption{Comparison of median best-of-run MAE on the test sets and total trial time. The best fitness results are highlighted. Significant improvements with respect to each method are denoted by ${a-h}$ according to the method labels. Significance is defined as $p < 0.05$ according to a pairwise Wilcoxon rank-sum test with Holm correction. The median total time to run 30 trials of each algorithm is shown on the right.}\label{tbl:bf_v}
\rowcolors{4}{white}{Gray}
\begin{tabularx}{\textwidth}{L{0.09\textwidth} R{0.084\textwidth} R{0.084\textwidth} R{0.084\textwidth} R{0.084\textwidth} R{0.084\textwidth} R{0.084\textwidth} | R{0.045\textwidth} R{0.15\textwidth}} \toprule
Method	&	Housing	&	Tower	&	Wind	&	ENH	&	ENC	&	UBall5D	&	Mean Rank	&	Median Total Trial Time (hr:min:s)	\\ \midrule
$^a$Rand Sel	&	${\tiny ^{}_{}}$ 0.469	&	${\tiny ^{}_{}}$ 0.458	&	${\tiny ^{}_{}}$ 0.463	&	${\tiny ^{}_{}}$ 0.288	&	${\tiny ^{}_{}}$ 0.272	&	${\tiny ^{d}_{}}$ 0.128	&	7.83	&	00:07:25	\\
$^b$Tourn	&	${\tiny ^{a}_{}}$ 0.408	&	${\tiny ^{a}_{}}$ 0.402	&	${\tiny ^{ad}_{}}$ 0.397	&	${\tiny ^{a}_{}}$ 0.207	&	${\tiny ^{a}_{}}$ 0.236	&	${\tiny ^{ad}_{}}$ 0.113	&	6.33	&	00:24:37	\\
$^c$AFP	&	${\tiny ^{abd}_{}}$ 0.354	&	${\tiny ^{abd}_{}}$ 0.319	&	${\tiny ^{abd}_{}}$ 0.381	&	${\tiny ^{abd}_{f}}$ 0.138	&	${\tiny ^{abd}_{f}}$ 0.171	&	${\tiny ^{abd}_{}}$ 0.094	&	4.00	&	01:09:18	\\
$^d$Lex	&	${\tiny ^{a}_{}}$ 0.402	&	${\tiny ^{ab}_{}}$ 0.355	&	${\tiny ^{a}_{}}$ 0.419	&	${\tiny ^{a}_{}}$ 0.210	&	${\tiny ^{a}_{}}$ 0.237	&	${\tiny ^{}_{}}$ 0.142	&	6.83	&	00:10:11	\\
$^e$Lex $\epsilon_y$	&	${\tiny ^{abd}_{f}}$ 0.325	&	${\tiny ^{abcd}_{}}$ 0.260	&	${\tiny ^{ad}_{}}$ 0.386	&	${\tiny ^{abcd}_{f}}$ 0.113	&	${\tiny ^{abcd}_{f}}$ 0.150	&	${\tiny ^{abcd}_{f}}$ 0.079	&	3.17	&	00:23:44	\\
$^f$Lex $\epsilon_e$	&	${\tiny ^{a}_{}}$ 0.386	&	${\tiny ^{abcd}_{}}$ 0.263	&	${\tiny ^{ad}_{}}$ 0.386	&	${\tiny ^{abd}_{}}$ 0.165	&	${\tiny ^{abd}_{}}$ 0.193	&	${\tiny ^{abcd}_{}}$ 0.082	&	4.50	&	00:20:24	\\
$^g$Lex $\epsilon_{y \lambda}$	&	${\tiny ^{abcd}_{f}}$ 0.321	&	${\tiny ^{abcd}_{f}}$ 0.239	&	\hl{ ${\tiny ^{abd}_{}}$ 0.378}	&	\hl{ ${\tiny ^{abcd}_{ef}}$ 0.101}	&	\hl{ ${\tiny ^{abcd}_{f}}$ 0.137}	&	${\tiny ^{abcd}_{}}$ 0.080	&	\hl{ 1.67}	&	00:29:26	\\
$^h$Lex $\epsilon_{e \lambda}$	&	\hl{ ${\tiny ^{abcd}_{f}}$ 0.309}	&	\hl{ ${\tiny ^{abcd}_{f}}$ 0.233}	&	${\tiny ^{abd}_{ef}}$ 0.381	&	${\tiny ^{abcd}_{f}}$ 0.106	&	${\tiny ^{abcd}_{f}}$ 0.141	&	\hl{ ${\tiny ^{abcd}_{fg}}$ 0.078}	&	\hl{ 1.67}	&	00:29:37	\\
\bottomrule
\end{tabularx}
\end{table*}
 \label{tbl_results}


\section{Discussion}\label{s:5}

$\epsilon$-lexicase selection is a global pool, uniform random sequence, non-elitist version of lexicase selection~\cite{spector_assessment_2012} that performs well on symbolic regression problems according to the experimental analysis presented in the last section. ``Global pool" refers to the fact that each selection event begins with the whole population (step 1 in \S\ref{s:2}). Smaller pool sizes have yet to be tried, but could potentially improve performance on certain problems that historically respond well to relaxed selection pressure. Pools could also be defined geographically~\cite{spector_trivial_2006}. ``Uniform random sequence" refers to the shuffling procedure for cases in step 2 (\S\ref{s:2}), and, as is the case with pool size, other orderings of test cases have yet to be reported in literature. One could consider biasing the ordering of cases in some ways that could select parents with certain desired properties. In~\cite{liskowski_comparison_2015}, Liskowski attempted to use derived objective clusters as cases in lexicase selection, but found that this actually decreased performance. Still, there may be a form of ordering or case reduction that improves lexicase selection's performance over random shuffling.   
 
The ordering of the test cases that produce a given parent also contains potentially useful information that could be used by the search operators in GP. Helmuth~\cite{helmuth_general_2015-1} observed that lexicase selection creates large numbers of distinct behavioral clusters in the population (an observation supported by Figure~\ref{fig:novelty_summary}). In that regard, it may be advantageous, for instance, to perform crossover on individuals selected by differing orders of cases such that their offspring are more likely to inherit subprograms with unique partial solutions to a given task. On the other hand, one could argue for pairing individuals based on similar selection cases, to promote niching and minimize the destructive nature of subtree crossover.    

\section{Conclusions}\label{s:6}
We find that $\epsilon$-lexicase selection, especially with automatic threshold adaptation ($\epsilon_{e \lambda}$ and $\epsilon_{y \lambda}$), performs the best on the regression problems studied here in comparison to the other GP methods studied. The performance in terms of test fitness is promising, as well as the measured wall clock times, which are comparable to tournament selection. In addition to introducing a non-elitist version of lexicase that defines test case pass conditions using an $\epsilon$ threshold, we demonstrated that $\epsilon$ can be set automatically based on the dispersion of error across the population on a test case. We observed that the definition of this threshold is insensitive to the elite error offset $e^*_t$. The results should motivate the use of $\epsilon$-lexicase selection as a parent selection technique for symbolic regression, and should motivate further research using non-elitist lexicase selection methods for continuous-valued problems in GP. 

\section{Acknowledgments}
The authors would like to thank Thomas Helmuth, Nic McPhee and Bill Tozier for their feedback as well as members of the Computational Intelligence Laboratory at Hampshire College. This work is partially supported by NSF Grant Nos. 1068864, 1129139 and 1331283. Any opinions, findings, and conclusions or recommendations expressed in this publication are those of the authors and do not necessarily reflect the views of the National Science Foundation. This work used the Extreme Science and Engineering Discovery Environment (XSEDE), which is supported by NSF grant number ACI-1053575~\cite{towns_xsede:_2014}.

%\bibliographystyle{abbrv}
%\bibliography{Lexicase_Regression_corrected}

\end{document}